\documentclass[runningheads]{llncs}
\usepackage{graphicx}
\usepackage[tight,footnotesize]{subfigure}
\usepackage{amsmath,amssymb} 
\usepackage{color}
\usepackage[width=122mm,left=12mm,paperwidth=146mm,height=193mm,top=12mm,paperheight=217mm]{geometry}
\begin{document}
\pagestyle{headings}
\mainmatter

\title{Person Re-Identification via Recurrent Feature Aggregation} 

\titlerunning{Person Re-Identification via Recurrent Feature Aggregation}

\authorrunning{Yichao Yan \emph{et al.}}

\author{Yichao Yan\inst{1} \and Bingbing Ni\inst{1}
Zhichao Song\inst{1} \and Chao Ma\inst{1} \and Yan Yan\inst{2} \and Xiaokang Yang\inst{1}}

\institute{Shanghai Jiao Tong University\\
\email{\{yanyichao,nibingbing,5110309394,chaoma,xkyang\}@sjtu.edu.cn}
\and
University of Michigan, Ann Arbor\\
\email{tom.yan.555@gmail.com}}


\maketitle

\begin{abstract}
We address the person re-identification problem by effectively exploiting a globally discriminative feature representation from a sequence of tracked human regions/patches. This is in contrast to previous person re-id works, which rely on either single frame based person to person patch matching, or graph based sequence to sequence matching. We show that a progressive/sequential fusion framework based on long short term memory (LSTM) network aggregates the frame-wise human region representation at each time stamp and yields a sequence level human feature representation. Since LSTM nodes can \emph{remember and propagate} previously accumulated good features and \emph{forget} newly input inferior ones,
even with simple hand-crafted features, the proposed recurrent feature aggregation network (\textbf{RFA-Net}) is effective in generating highly discriminative sequence level human representations.
Extensive experimental results on two person re-identification benchmarks demonstrate that the proposed method performs favorably against state-of-the-art person re-identification methods. Our code is available at {https://sites.google.com/site/yanyichao91sjtu/}
\end{abstract}

\section{Introduction}

Person re-identification (re-id) deals with the problem of re-associating a specific person across non-overlapping cameras. It has been receiving increasing popularity~\cite{DBLP:journals/ivc/Bedagkar-GalaS14} due to its important applications in intelligent video surveillance.

Existing methods mainly focus on addressing the single-shot person re-id problem. Given a probe image of one person taken from one camera, a typical scenario for single-shot person re-id is to identify this person in a set of gallery images taken from another camera. Usually, the identification results are based on ranking the similarities of the probe-gallery pairs. The performance of person re-id is measured by the rank-$k$ matching rate if the correct pair hits the retrieved top-$k$ ranking list. To increase the matching rate, state-of-the-art approaches either employ discriminative features in representing persons or apply distance metric learning methods to increase the similarity between matched image pairs. Numerous types of features have been explored to represent persons, including global features like color and texture histograms~\cite{DBLP:conf/cvpr/FarenzenaBPMC10,DBLP:conf/eccv/GrayT08}, local features such as SIFT~\cite{DBLP:journals/ijcv/Lowe04} and LBP~\cite{DBLP:journals/pami/OjalaPM02}, and deep convolutional neural network (CNN) features~\cite{DBLP:conf/cvpr/LiZXW14,DBLP:conf/cvpr/AhmedJM15}. In the meantime, a large number of metric learning approaches have been applied to person re-id task, such as LMNN~\cite{DBLP:conf/nips/WeinbergerBS05}, Mahalanobis distance metric~\cite{DBLP:series/acvpr/RothHKBB14}, and RankSVM~\cite{DBLP:journals/ir/ChapelleK10}.
Despite the significant progress in recent years, the performance achieved by these methods do not fulfill the real-application requirement due to the following reasons. First, images captured by different cameras undergo large amount of appearance variations caused by illumination changes, heavy occlusion, background clutter, or human pose changes. More importantly, for real surveillance, persons always appear in a video rather than in a single-shot image. These single-shot based methods fail to make full use of the temporal sequence information in surveillance videos.

Several algorithms have been proposed to tackle the multi-shot person re-id problem, \textit{i.e.}, to match human instances in the sequence level (these human patch sequences are usually obtained by visual detection and tracking). To exploit the richer sequence information, existing methods mainly resort to: (1) key frame/fragment representation~\cite{DBLP:conf/eccv/WangGZW14}; (2) feature fusion/encoding~\cite{DBLP:conf/iccv/ZhengSTWWT15,DBLP:journals/corr/ZhengSTWBT15,DBLP:conf/iccv/KaranamLR15} and (3) spatio-temporal appearance model~\cite{DBLP:conf/iccv/LiuMZH15}. Despite their favorable performance on recent benchmark datasets~\cite{DBLP:conf/eccv/WangGZW14,DBLP:conf/scia/HirzerBRB11}, we have the following observations on their limitations. First, key frame/fragment representation based algorithms~\cite{DBLP:conf/eccv/WangGZW14} often assume the  discriminative fragments to be located in the local minima and maxima of Flow Energy Profile~\cite{DBLP:conf/eccv/WangGZW14}, which may not be accurate. During the final matching, since only one fragment is selected to represent the whole sequence, richer information contained by the rest of the sequences is not fully utilized. Second, feature fusion/encoding methods~\cite{DBLP:conf/iccv/ZhengSTWWT15,DBLP:journals/corr/ZhengSTWBT15,DBLP:conf/iccv/KaranamLR15} take bag-of-words approach to encode a set of frame-wise features into a global vector, but ignore the informative spatio-temporal information of human sequence. To overcome these shortages, the recently proposed method~\cite{DBLP:conf/iccv/LiuMZH15} employs the spatially and temporally aligned appearance of the person in a walking cycle for matching. However, such approach is extremely computationally inefficient and thus inappropriate for real applications. It is therefore of great importance to explore a more effective and efficient scheme to make full use of the richer sequence information for person re-id.

\begin{figure}[t]
\centering
\includegraphics[width=4.5in]{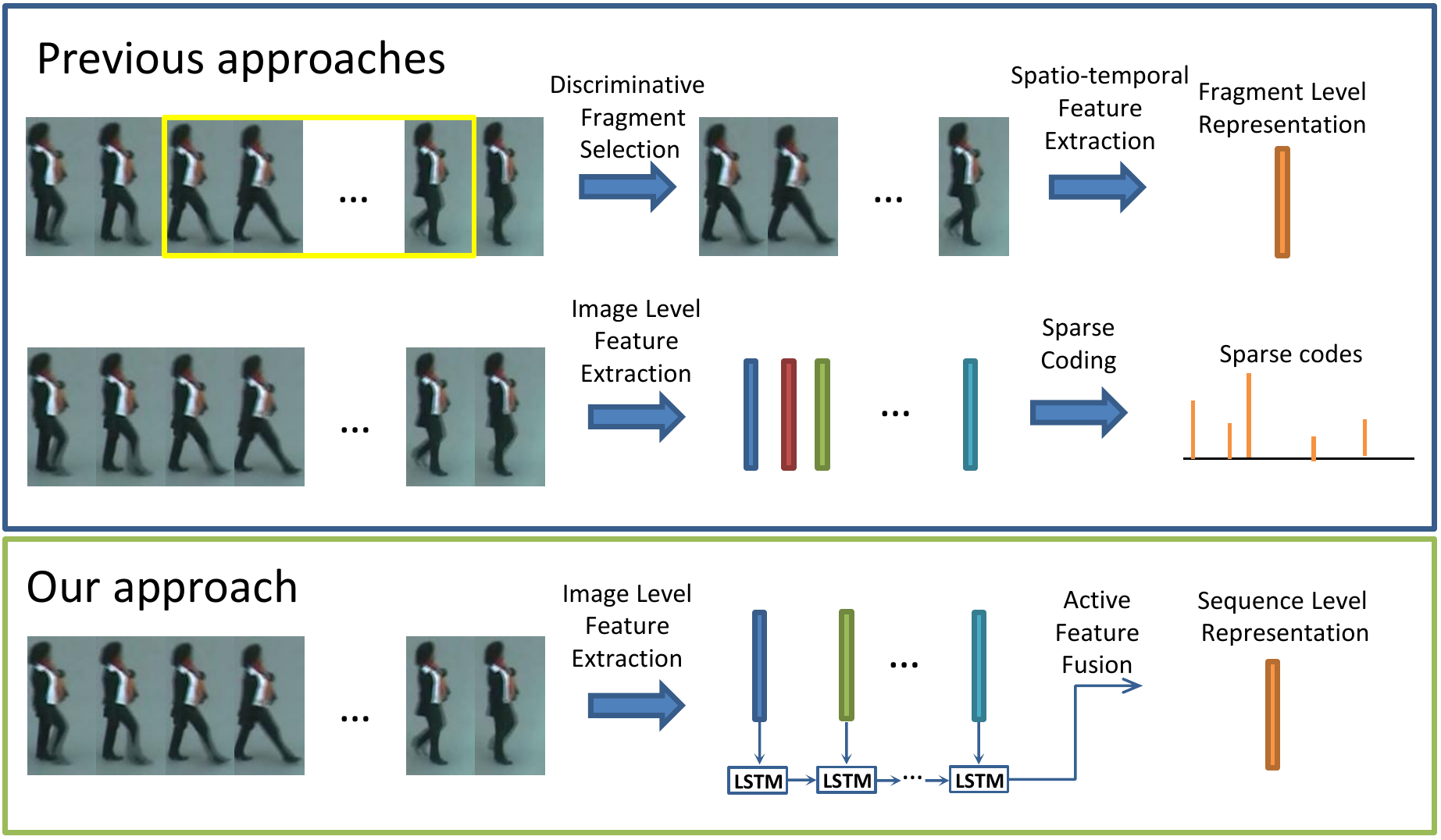}
\caption{Pipeline of the proposed recurrent feature aggregation network for multi-shot person re-id with comparison to previous methods. The top part depicts two common used methods for multi-shot person re-id, \textit{i.e.}, to choose a most discriminative fragment for matching, or to use sparse coding method to enhance feature representation. The bottom part gives an illustration of our approach, frame-wise features are input into a feature aggregation (LSTM) network, and the output vector of the network builds the sequence level representation}
\label{fig:fig1}
\end{figure}

The fundamental challenge of multi-shot person re-id is how to systematically aggregate both the frame-wise appearance information as well as temporal dynamics information along the human sequence to generate a more discriminative sequence level human representation. 
To this end, we propose a recurrent feature aggregation network (\textbf{RFA-Net}) that builds a sequence level representation from a temporally ordered sequence of frame-wise features, based on a typical version of recurrent neural network, namely, Long Short-Term Memory network~\cite{DBLP:journals/neco/HochreiterS97}. Figure~\ref{fig:fig1} shows an overview of the difference between our method and previous approaches. The proposed feature aggregation framework possesses various advantages. First, it allows discriminative information of frame-wise data to propagate along the temporal direction, and discriminative information could be accumulated from the first LSTM node to the deepest one, thus yielding a highly discriminative sequence level human representation. Second, during feature propagation, this framework can \emph{prevent} non-informative information from reaching the deep nodes, therefore it is robust to noisy features (due to occlusion, tracking/detection failure, or background clutter etc). Third, the proposed fusion network is simple yet efficient, which is able to deal with sequences with variable length.

The main contributions of this work lie in that we present a recurrent feature aggregation network to address the multi-shot person re-id problem. The proposed network jointly takes the feature fusion and spatio-temporal appearance model into account and makes full use of the temporal sequence information for multi-shot person re-id. With the use of the proposed network, hand-crafted low-level features can be augmented with temporal cues and significantly improve the accuracy  of person re-id. Extensive experiments on two publicly available benchmark datasets demonstrate that the proposed person re-id method performs favorably against state-of-the-arts algorithms in terms of effectiveness and efficiency.

\section{Related work}

Person re-identification has been explored for several years. A large body of works mainly focus on solving two subproblems: feature representation and distance metric learning.

Low-level features such as texture, color and gradient are most commonly used for person representation, however, these features are not powerful enough to discriminate a person from similar ones.
Some methods address this problem by combining features together to build a more discriminative representation~\cite{DBLP:conf/bmvc/MaSJ12} or by selecting the most discriminative features to represent a person~\cite{DBLP:conf/cvpr/FarenzenaBPMC10}.

When multiple images are available for one person, image level features are accumulated or averaged~\cite{DBLP:conf/iccv/KaranamLR15} to build a more robust representation. When image sequences or videos are available, the space-time information can be used to build space-time representations. These kind of features such as 3D SIFT~\cite{DBLP:conf/mm/ScovannerAS07} and 3D HOG~\cite{DBLP:conf/bmvc/KlaserMS08} are simply extensions of 2D features, which are not powerful enough as in the single-shot case. Other methods try to build more powerful descriptors, such as in~\cite{DBLP:conf/icdsc/HamdounMSS08}, interest points descriptors are accumulated along the time space to capture appearance variability. Different color-based features combined with a graph based approach was proposed in~\cite{DBLP:conf/iciap/CongAKD09}. This method tries to learn the global geometric structure of the people and to realize comparison of the video sequences. Many other methods also exploit image sequences to enhance feature description~\cite{DBLP:conf/cvpr/GheissariSH06,DBLP:conf/iccv/XuLZL13}.
Wang \emph{et al.}~\cite{DBLP:conf/eccv/WangGZW14} presented a model to automatically select the most discriminative video fragments and simultaneously to learn a video ranking function for person re-id.
Karanam \emph{et al.}~\cite{DBLP:conf/iccv/KaranamLR15} proposed a Local Discriminant Analysis approach that is capable of encoding a set of frame-wise features representing a person into a global vector.
Liu \emph{et al.} recently proposed a spatio-temporal appearance representation method~\cite{DBLP:conf/iccv/LiuMZH15}, and feature vectors that encode the spatially and temporally aligned appearance of the person in a walking cycle are extracted.
These methods are usually feature-sensitive, designing complex/good features or combining several features together may achieve good performance, while only using simple features would lead to significant performance drop~\cite{DBLP:conf/eccv/WangGZW14}. In contrast to these methods, the method we proposed is simple and effective, \textit{i.e.}, it could learn to aggregate very simple features (LBP and color information) into a highly discriminative representation.

A large number of metric learning as well as ranking algorithms have been proposed to solve the person re-id problem. Mahalanobis distance metric~\cite{DBLP:series/acvpr/RothHKBB14} is most commonly used in person re-id. Boosting was applied to Mahalanobis distance learning by Shen \emph{et al}~\cite{DBLP:journals/jmlr/ShenKWH12}. Person re-id was formulated  as a learning-to-rank problem in~\cite{DBLP:conf/bmvc/ProsserZGX10}. Zheng \emph{et al } have also formulated person re-id as a relative distance comparison (RDC) problem~\cite{DBLP:journals/pami/ZhengGX13}. Although an efficient distance metric learning algorithm can model the transition in feature space between two camera views and thus enhance the re-identification performance, it will be no longer necessary if the features possess high discriminativeness. As demonstrated by our experiments, our feature representation will still generate promising results when only a simple metric (cosine distance) is applied.

Deep learning based methods~\cite{DBLP:conf/cvpr/LiZXW14,DBLP:conf/cvpr/AhmedJM15} have also been applied to person re-id to simultaneously solve feature representation and metric learning problem.  A typical structure of this approach consists of two parts. The first part is a feature learning network, usually a convolutional neural network or other kinds of networks. The feature pairs extracted from the first step are then fed into several metric learning layers to predict weather a pair of images belong to the same person. Usually the networks have simple structure and can be trained end-to-end. However, this kind of approaches use image pairs to learn feature representation and distance metric, which is unable to consider the space-time information in video sequences, this makes it more suitable for single-shot problem. Moreover, deep learning based methods are easy to overfit on small database. In contrast, our method adopts a simple LSTM network to learn space-time information, which is more efficient than the deep learning based models. We also realize two contemporary works~\cite{McLaughlin_2016_CVPR,Haque_2016_CVPR}.

\section{Sequential Feature Aggregation for Person Re-Identification}

\subsection{Overview}
As mentioned above, previous single-shot based re-id methods rely on the information extracted from a single image, which is insufficient to identify the person in a gallery of images. For multi-shot person re-id, on one hand, methods based on graph matching to calculate the similarity between sequences are computationally complex and sensitive to outliers (e.g., false human detections). On the other hand, methods  based on pooling frame-level features into a sequence level representation usually cannot well model the temporal change of human appearance, e.g., human dynamics. Therefore, in this work, we aim to aggregate both image level features and human dynamics information into a discriminative sequence level representation. More specifically, our sequential feature aggregation scheme is based on a recurrent neural network (e.g., LSTMs). Frame-wise human features are sequentially fed into this network and discriminative/descriptive information is propagated and accumulated through the LSTM nodes, and yielding final sequence level human representation. Accordingly, both the frame-wise human appearance information and human dynamics information are properly encoded.

\subsection{Sequential Feature Aggregation Network}
\subsubsection{Network Architecture}
\begin{figure}[t]
\centering
\includegraphics[width=4in]{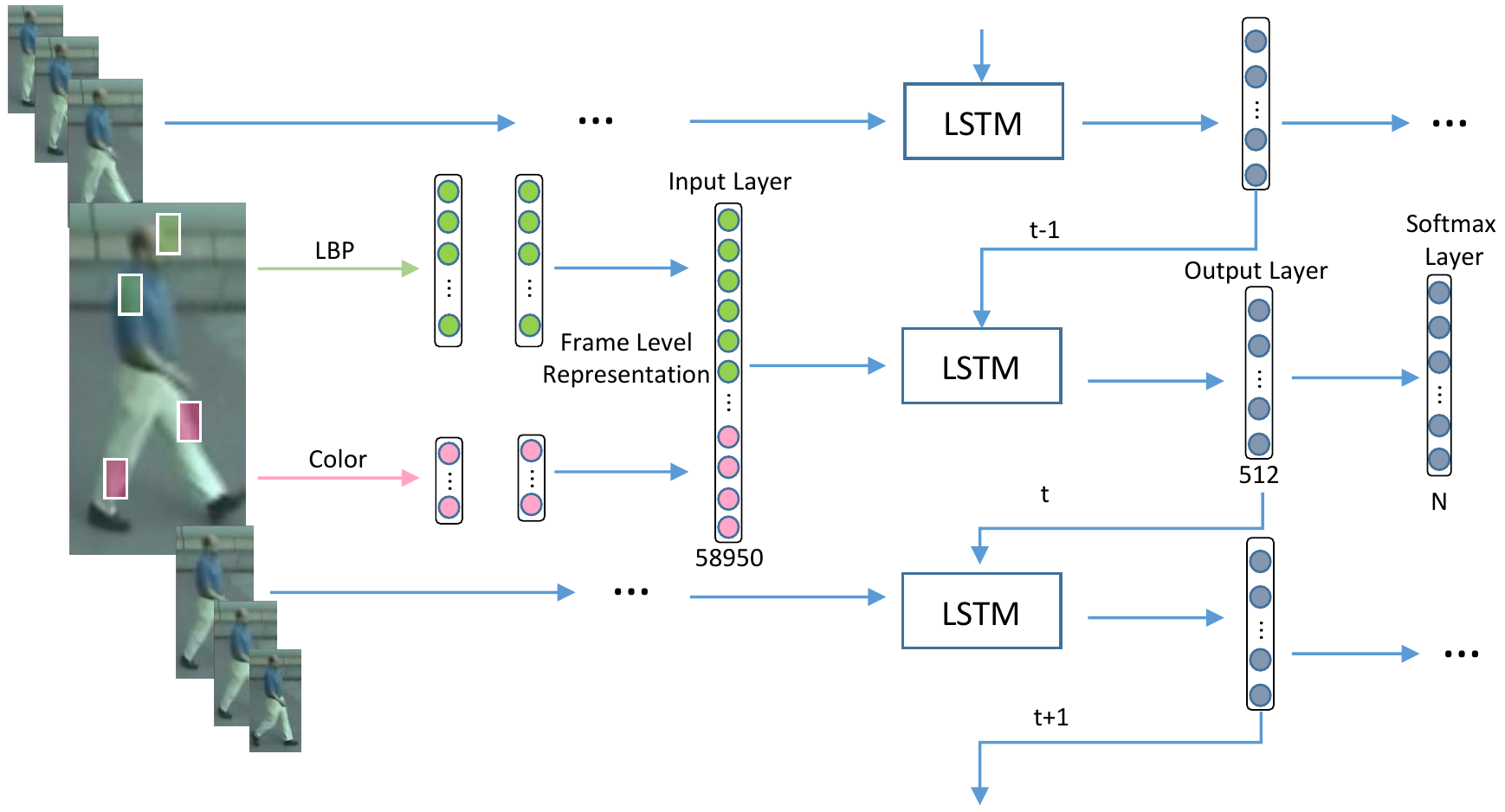}
\caption{Details of the network structure. LBP and color features are first extracted from overlapping rectangular image patches, and then concatenated as frame level representation. This representation as well as the previous LSTM outputs are input to the current LSTM node, the output of the LSTM is further connected to a softmax loss layer for error calculation and gradient backpropagation}
\label{fig:detail}
\end{figure}

Motivated by the success of recurrent neural networks (RNNs) in temporal sequence analysis, we employ a LSTM network as our feature aggregation network prototype. A LSTM network maintains the information of the previous time stamp and uses it to update the current state sequentially, therefore it is a powerful tool to deal with sequential data. With a LSTM network, information could be propagated from the first LSTM node to the last one, and this information aggregation mechanism is very useful for our given task, \textit{i.e.}, to fuse the frame-wise features to a sequence level feature representation. In other words, the output of the last LSTM nodes contains aggregated discriminative information of the sequence. Each LSTM node of the feature aggregation network contains input nodes, hidden nodes and output nodes, as illustrated in Figure~\ref{fig:detail}. The input features, denoted as $\{\mathbf{x}_t\}_{t=1:T}$, could be either deep CNN features or simply hand crafted features. In this work, we use simple hand crafted features. This is because although deep CNN features are more prevailing recently, to train a valid CNN model requires large amount of data. Unfortunately, existing person re-id databases are mostly with small size, and the learned deep CNN features inevitably suffer from over-fitting. The dimensionality of the input feature vector of each time stamp is $58950$. Although in many other applications, stack of LSTM nodes are used for better network representation capability, our proposed network employs only one layer of LSTM nodes (\textit{i.e.}, no stack) with $512$ hidden units, as our experimental results show that it is sufficient for our re-id task and deeper models or more hidden units do not bring too much performance gain.

\subsubsection{Frame Level Representation}
For each human bounding box, we use both color and texture features to represent human appearance features, which have proven to be effective for the task of person re-id~\cite{DBLP:conf/bmvc/MaSJ12,DBLP:conf/cvpr/ZhaoOW13}. These features are naturally complementary to each other, because texture features are usually extracted from gray-scale images, while color features could capture the chromatic patterns within an image. Particularly, we use Local Binary Patterns (LBP)~\cite{DBLP:journals/pami/OjalaPM02},  HSV and Lab color channels to build an image frame level person representation, as have been used by many previous works~\cite{DBLP:conf/eccv/HirzerRKB12,DBLP:conf/cvpr/JingZWYLYHX15}. All images are resized to the size of $128\times64$ before feature extraction. Features are extracted from $16\times8$ rectangular patches, with a spatial overlap of $8$ and $4$ pixels, vertically and horizontally. For each patch, we compute the histogram of LBP codes from gray scale representation. The mean values of HSV and Lab color channels are also computed. All these patch level features are concatenated to build an image frame level representation, \textit{i.e.}, for each rectangular patch, the dimension is 262 (256+6), and the dimensionality of the input feature is 58950 (262*225) for each time stamp. The output of the LSTM unit at each time stamp is a 512-dimensional vector.

\subsubsection{Network Training}
The sequence of image level feature vectors are input to a LSTM network for progressive/sequential feature fusion. The network is trained as a classification problem of $N$ classes ($N$ is the number of persons), \textit{i.e.}, we aim to classify the feature vectors which belong to the same person into the same class. In particular, each LSTM node includes three gates, (\textit{i.e.} the input gate $\mathbf{i}$, the output gate $\mathbf{o}$ and the forget gate $\mathbf{f}$) as well as a memory cell. At each time stamp $t$ (in our case, $t$ indicates the order in the sequence), given the input $\mathbf{x}_t$ and the previous hidden state $\mathbf{h}_{t-1}$, we update the LSTM network as follows:
\begin{flalign}
\hspace{25mm}
{\mathbf{i}_t} &= \sigma({\mathbf{W}_i}{\mathbf{x}_t} + {\mathbf{U}_i}{\mathbf{h}_{t-1}}+{\mathbf{V}_i}{\mathbf{c}_{t-1}} + \mathbf{b}_i)&\\
{\mathbf{f}_t} &= \sigma({\mathbf{W}_f}{\mathbf{x}_t} + {\mathbf{U}_f}{\mathbf{h}_{t-1}}+{\mathbf{V}_f}{\mathbf{c}_{t-1}} + \mathbf{b}_f)&\\
{\mathbf{c}_t} &= \mathbf{f}_t\cdot{\mathbf{c}_{t-1}} + \mathbf{i}_t\cdot{\tanh({\mathbf{W}_c}{\mathbf{x}_t} + {\mathbf{U}_c}{\mathbf{h}_{t-1}} + \mathbf{b}_c)}&\\
{\mathbf{o}_t} &= \sigma({\mathbf{W}_o}{\mathbf{x}_t} + {\mathbf{U}_o}{\mathbf{h}_{t-1}}+{\mathbf{V}_o}{\mathbf{c}_{t}} + \mathbf{b}_o)&\\
{\mathbf{h}_t} &= \mathbf{o}_t\cdot{\tanh({\mathbf{c}_t})}
\end{flalign}
where $\sigma$ is the sigmoid function and $\cdot$ denotes the element-wise multiplication operator. $\mathbf{W}_{*}$, $\mathbf{U}_{*}$ and $\mathbf{V}_{*}$ are the weight matrices, and $\mathbf{b}_{*}$ are the bias vectors. The memory cell $\mathbf{c}_t$ is a weighted sum of the previous memory cell $\mathbf{c}_{t-1}$ and a function of the current input. The weights are the activations of forget gate and input gate respectively. Therefore, intuitively, the network can learn to propagate/accumulate the good features to the deeper nodes of the network and forget noisy features in the meantime. The output of the LSTM hidden state $\mathbf{h}_t$ serves as the fused image level feature representation vector, which is further connected to a softmax layer. The output of the $N$-way softmax is the prediction of the probability distribution over $N$ different identities.
\begin{equation}
{y_i} = \frac{\exp(y'_i)}{\sum_{k=1}^{N}{\exp({y'_k})}},
\end{equation}
where $y'_j=\mathbf{h}_{t}\cdot \mathbf{w}_{j}+{b}_j$ linearly combines the LSTM outputs. The network is learned by minimizing $-\log{{y}_k}$, where $k$ is the index of the true label for a given input. Stochastic gradient descent is used with gradients calculated by back-propagation.

As only two sequences are available for a single person, and the length of the sequences are variable. To increase training instances and to make the model applicable for sequences of variable length, we randomly extract subsequences of fixed length $L$ for training. $L$ can be determined according to the nature of the dataset, we fix $L = 10$ in our experiments, considering the tradeoff between the amount of training instances and the information contained within each subsequence.

\subsubsection{Network Implementation Details}
In the proposed framework, the LSTM networks are trained/tested using the implementation of Donahue \emph{et al.}~\cite{lrcn2014} based on \emph{caffe}~\cite{jia2014caffe}. The LSTM layer contains $512$ output units, and a drop-out layer is placed after it to avoid over-fitting. The weights of the LSTM layer is initialized from a uniform distribution in $[-0.01, 0.01]$. The training phase lasts for 400 epoches. As mentioned above, there is only a single LSTM layer to learn, therefore, we start with small learning rate $0.001$ and after $200$ epoches it decreases to $0.0001$. The training procedure will terminate after another $200$ epoches. We train the network on a Titan X GPU card, and a whole training duration is about 8 hours for a dataset of 400 image sequences.

\subsection{Person Re-identification}
Once the feature aggregation network is trained, the sequence level representation can be obtained as follows. First, the testing sequence of a human patches is input to the aggregation network, then the LSTM outputs of each time stamp are concatenated to form the sequence level human feature representation for re-id purpose. The LSTM network we trained contains $10$ time stamps, and from each time stamp we can extract a $512$-dimensional feature vector. Therefore the total length of the sequence level representation is $5120 (512\times10)$, denoted as $\mathbf{s}_i$ for the $i$-th human sequence sample. In our test phase, since the lengths of the human sequences are varying, the representations of $K$ randomly selected subsequences (we assume the length of the sequence is larger than $10$) are averaged to build the final sequence level representation. There are two benefits from this scheme. First, noisy information can be further diluted while discriminative information can be retained. Second, influences of different poses and walking circles are decreased.

After we get the person representation, we can apply a simple metric or distance metric learning methods to measure the similarity of two input instances. In this paper, we employ the RankSVM~\cite{DBLP:journals/ir/ChapelleK10} method to measure the similarity as in~\cite{DBLP:conf/bmvc/ProsserZGX10}. In the meantime, a non supervised method (cosine distance metric) is also applied as a baseline metric. Experiments show that this simple combination works well in our re-id task. Consider the sequence level features ${\{({\mathbf{s}_i^a},{\mathbf{s}_i^b})}\}_{{i=1}}^{N}$, where $\mathbf{s}_i^a$ and $\mathbf{s}_i^b$ denote the sequence level features of the $i$-th person from camera $a$ and $b$. Suppose images from camera $a$ are probe images and images from camera $b$ are galleries. The cosine distance of a probe-gallery pair is represented as
\begin{equation}
{d_{ij}} = \frac{{\mathbf{s}_i^a} \cdot {\mathbf{s}_j^b}}{\left\|{\mathbf{s}_i^a}\right\|\left\|{\mathbf{s}_j^b}\right\|},
\end{equation}
where the $\left\|\cdot\right\|$ denotes the $L_2$ norm of the vector. Note that we do not use any complex metric learning technique. For a probe $\mathbf{s}_i^a$, the distances of all the pairs ${\{d_{ij}}\}_{{i=1}}^{N}$ are ranked in ascending order. If the distance of the true probe-gallery pair is ranked within the $k$ top matched samples, it is counted in the top-$k$ matching rate.

To train the RankSVM model, probe-gallery pairs with ground-truth labels are used for training. There is a single positive instance for a person, denoted as $\mathbf{s}^+_i = |\mathbf{s}_i^a - \mathbf{s}_i^b|$. The negative instances consist of non-matching pairs: $\mathbf{s}^-_{ij} = |\mathbf{s}_i^a - \mathbf{s}_j^b|$, for all $j\neq i$. Consider a linear function $F(\mathbf{s})={\mathbf{w}}\cdot{\mathbf{s}}$, we wish to learn a weight vector such that:
\begin{equation}
\forall j\neq i: F(\mathbf{s}^+_i) > F(\mathbf{s}^-_{ij}).
\end{equation}
The solution can be obtained using the SVM solver by minimizing following objective function:
\begin{equation}
\begin{aligned}
\frac{1}{2}{\mathbf{w}}\cdot{\mathbf{w}} &+ C\sum{{\xi}_{ij}}\\
s.t.\ \forall j\neq i: {\mathbf{w}}\cdot{\mathbf{s}^+_i}&\ge{\mathbf{w}}\cdot{\mathbf{s}^-_{ij}}+1-{\xi}_{ij}\\
\forall{(i,j)}&: {\xi}_{ij}\ge{0}.
\end{aligned}
\end{equation}
Once the RankSVM model is trained, it can be used to give the similarity rankings out of a set of probe-gallery pairs, \textit{i.e.}, to fullfill the re-id task.

\section{Experiments}
In this section, we present extensive experimental evaluations and in-depth analysis of the proposed method in terms of effectiveness and robustness. We also quantitatively compare our method with the state-of-the-art multi-shot re-id methods. Algorithmic evaluations are performed on the following two person re-id benchmarks:

\noindent
\textbf{iLIDS-VID dataset.} The iLIDS-VID dataset contains of $600$ image sequences for $300$ people in two non-overlapping camera views. Each image sequence consists of frames of variable length from $23$ to $192$, with an average length of $73$. This dataset was created at an airport arrival hall under a multi-camera CCTV network, and the images were captured with significant background clutter, occlusions, and viewpoint/illumination variations, which makes the dataset very challenging.

\noindent
\textbf{PRID 2011 dataset.} The PRID 2011 dataset includes $400$ image sequences for $200$ people from two camera. Each image sequence has a variable length consisting of $5$ to $675$ image frames, with an average number of $100$. The images were captured in an uncrowded outdoor environment with relatively simple and clean background and rare occlusion; however, there are significant viewpoint and illumination variations as well as color inconsistency between the two views. Following~\cite{DBLP:conf/eccv/WangGZW14}, the sequence pairs with more than $21$ frames are used in our experiments.

\noindent
\textbf{Evaluation settings.} Following~\cite{DBLP:conf/eccv/WangGZW14}, the whole set of human sequence pairs of both datasets is randomly split into two subsets with equal size, \textit{i.e.}, one for training and the other for testing. The sequence of the first camera is used as probe set while the gallery set is from the other one. For both datasets, we report the performance of the average Cumulative Matching Characteristics (CMC) curves over $10$ trials.

\subsection{Effectiveness of Recurrent Feature Aggregation Network}
\begin{figure}[t]
  \centering
  \subfigure[]{
    \label{fig:subfig:a} 
    \includegraphics[width=2in, height=1.7in]{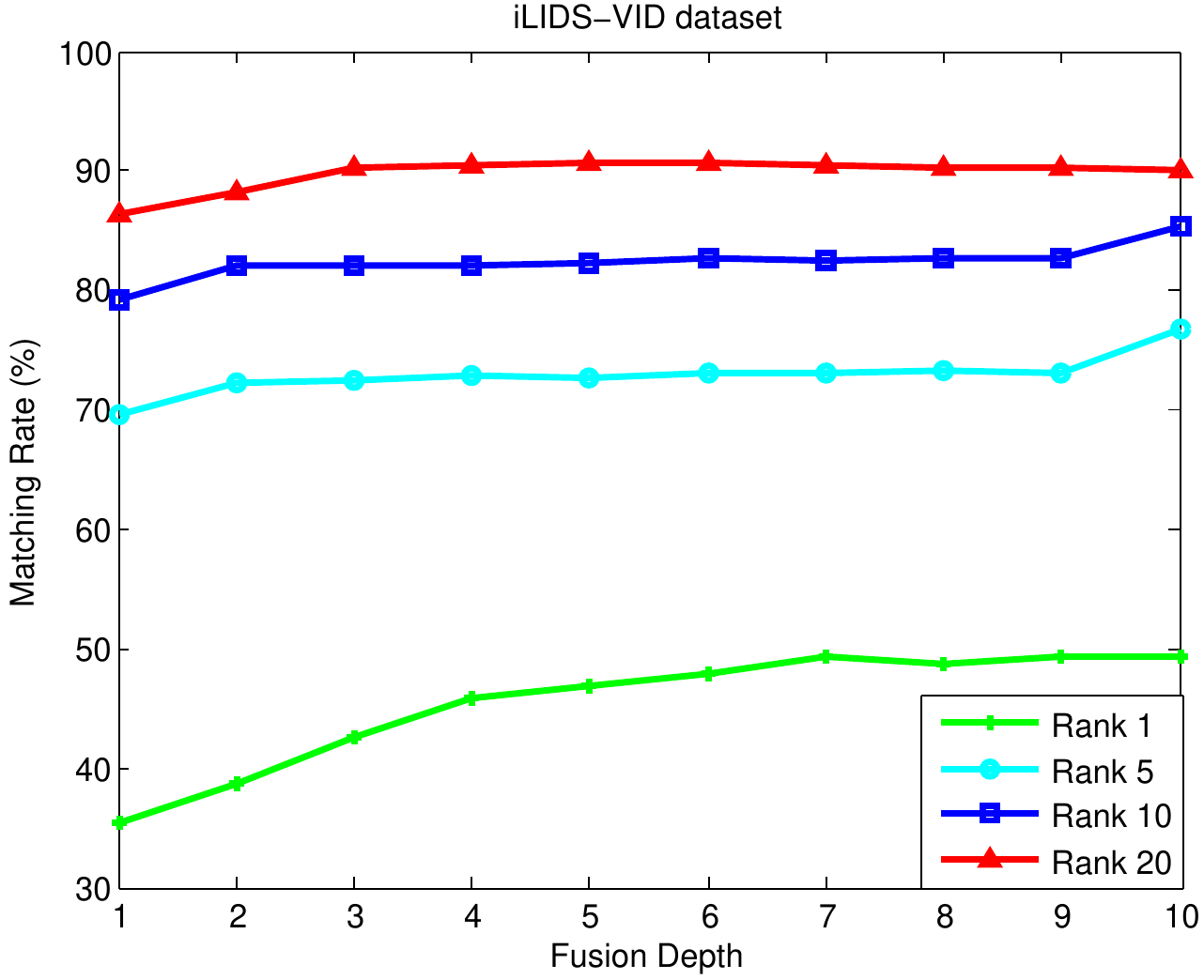}}
  \subfigure[]{
    \label{fig:subfig:b} 
    \includegraphics[width=2in, height=1.7in]{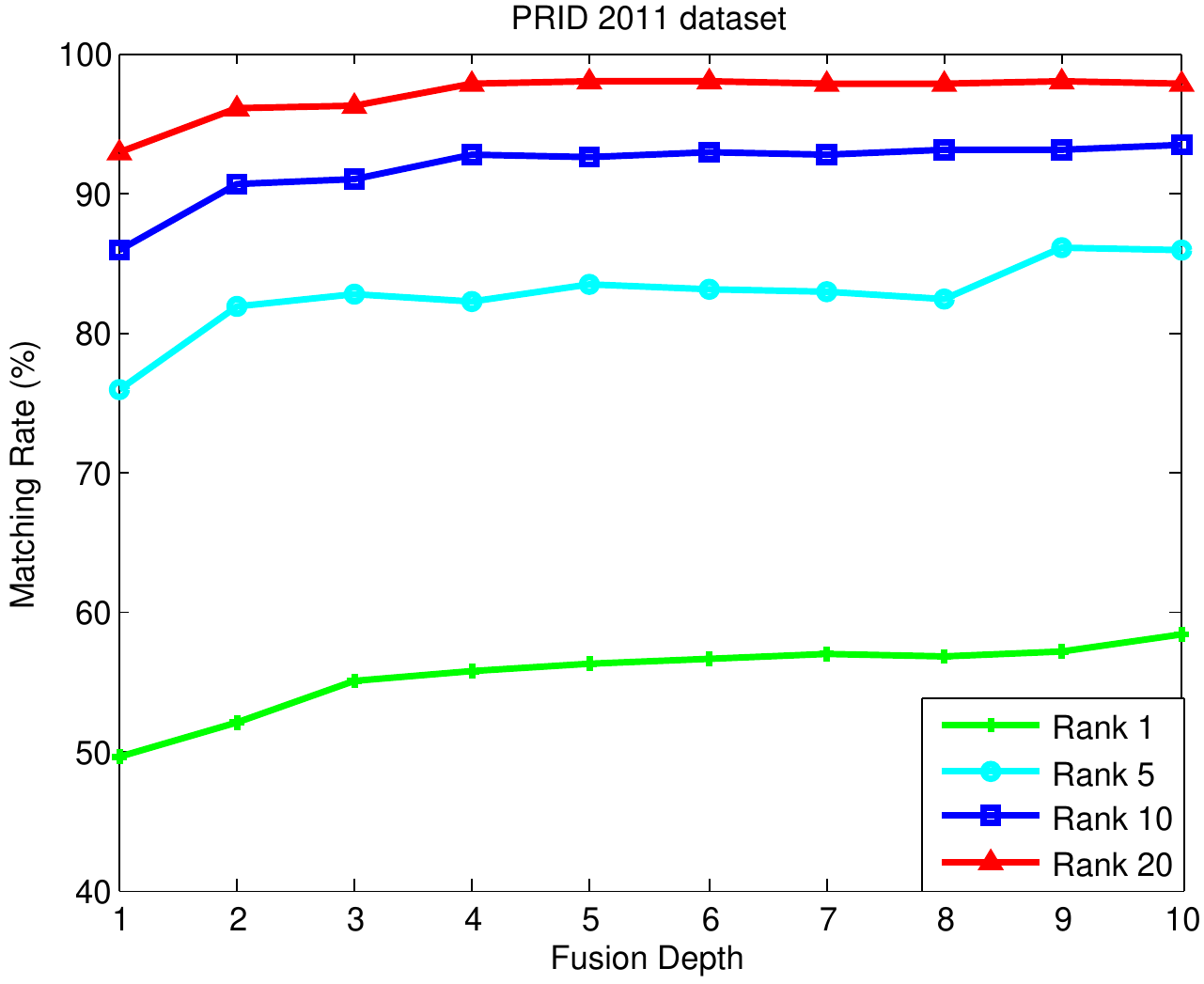}}
  \caption{Rank-1,5,10 and 20 person re-id matching rates based on different \emph{fusion depth} on (a) iLIDS-VID and (b) PRID 2011 datasets}
  \label{fig:effectiveness} 
\end{figure}

To evaluate the effectiveness of the proposed recurrent feature aggregation network (\textbf{RFA-Net}), we extract the learned feature representation from each LSTM node of our network and evaluate the person re-id matching rate on the testing set. The results on both datasets are illustrated in Figure~\ref{fig:effectiveness}. From Figure~\ref{fig:effectiveness}, we have two observations. First, the matching rate increases consistently when the features are extracted from \emph{deeper} LSTM nodes of our network. And the accumulated sequence level person representation (last node output) outperforms single frame level representation (first node output) for about $10\%$ on both datasets. This demonstrates that the proposed network is able to aggregate discriminative information from frame level towards a more discriminative sequence level feature representation. Second, we note that the rank-$1$ matching rate increases (with respect to the network depth) faster than others, which shows that the rank-$1$ matching performance relies more on the feature representation.

Based on the same type of visual features (\textit{i.e.}, Color\&LBP), we further compare our method with state-of-the-art matching method for person re-id. The compared methods are as follows.
1) Baseline method: the averagely pooled feature representation (\textit{i.e.}, Color\&LBP) of each frame in the person sequence combined with RankSVM~\cite{DBLP:journals/ir/ChapelleK10} based matching method;
2) Sequence matching method: the two person sequences are matched by the Dynamic Time Wrapping (DTW)~\cite{DBLP:conf/eccv/SimonnetLVOT12} method;
and 3) Discriminative feature selection method: the Discriminative Video Ranking (DVR)~\cite{DBLP:conf/eccv/WangGZW14} method which selects the most discriminative fragments for person matching.
The comparison results on both datasets are summarized in Table~\ref{table1}. For our method, we perform experiment by using the simple cosine distance metric or the learned distance based on RankSVM and report both results. Note that for the rank-$1$ matching rate, our method outperforms the baseline method and the Discriminative Video Ranking (DVR) based method for more than $60\%$. We further notice that even without metric learning, our method still achieves a rather good performance, which well demonstrates the discriminativeness of our learned sequence level representation, \textit{i.e.}, our propose network is capable of fusing frame level human features into a more discriminative human sequence level feature representation. Also, based on recurrent scheme, our network implicitly encodes human dynamics. In contrast, the comparing methods which are only based on frame level feature matching (or even discriminative feature selection) do not possess good discriminative capability.
\begin{table}[t]
\centering
\caption{Performance of different methods based on Color\&LBP feature}
\label{table1}
\begin{tabular*}{1\textwidth}{@{\extracolsep{\fill} }c|cccc|cccc}
\hline
Dataset                & \multicolumn{4}{c|}{iLIDS-VID} & \multicolumn{4}{c}{PRID 2011} \\ \hline
Rank R                 & R=1    & R=5   & R=10  & R=20  & R=1    & R=5   & R=10  & R=20  \\ \hline
Color\&LBP~\cite{DBLP:conf/eccv/HirzerRKB12}+RSVM       & 23.2   & 44.2  & 54.1  & 68.8  & 34.3   & 56.0    & 65.5  & 77.3  \\
Color\&LBP+DTW~\cite{DBLP:conf/eccv/SimonnetLVOT12}       & 9.3   & 21.7  & 29.5  & 43  & 14.6   & 33    & 42.6  & 47.8  \\
Color\&LBP+DVR~\cite{DBLP:conf/eccv/WangGZW14}       & 34.5   & 56.7  & 67.5  & 77.5  & 37.6   & 63.9    & 75.3  & 89.4  \\ \hline
Color\&LBP+\textbf{RFA-Net}+Cosine &44.5  &71.9   &82.0   &\textbf{90.1}    & 54.9    &84.2  &\textbf{93.7}   &\textbf{98.4}  \\
Color\&LBP+\textbf{RFA-Net}+RSVM   &\textbf{49.3}  &\textbf{76.8}  &\textbf{85.3}  &90.0 &\textbf{58.2} &\textbf{85.8} &93.4 &97.9\\ \hline
\end{tabular*}

\end{table}
\subsection{Robustness of Recurrent Feature Aggregation Network}
As there exist a large amount of color/illumination change as well as occlusion and background clutter, a good re-id method should be robust to these noise. Here, we evaluate the robustness of our method by adding noise to human patch sequences to be matched. For the test image sequences in each dataset, part of the images are replaced by noise images, which are randomly selected from the other dataset (e.g., the noise for PRID 2011 dataset are images from iLIDS-VID dataset). Then we extract feature vectors from the noise contaminated images for person re-id. As shown in Table~\ref{table2}, although the performance suffers from the noise, our method still achieves $29.8\%$ and $44.7\%$ rank-1 matching rate on the two datasets when up to $50\%$ images (frames) are contaminated, which is remarkable. We also notice that the rank-10 and rank-20 matching rates just decree slightly compared to the corresponding noise-free sequences. This demonstrates that despite large amount of noises, our feature representation still remains good discriminativeness. This is because that LSTM network is capable of \emph{propagating} discriminative features to deeper LSTM nodes as well as \emph{forgetting} irrelevant information (\textit{i.e.}, noises) during propagation. Therefore the aggregated sequence level representation is highly discriminative.
\begin{table}[]
\centering
\caption{Performance of our method in existence of noises}
\label{table2}
\begin{tabular*}{1\textwidth}{@{\extracolsep{\fill} }c|cccc|cccc}
\hline
Dataset                & \multicolumn{4}{c|}{iLIDS-VID} & \multicolumn{4}{c}{PRID 2011} \\ \hline
Rank R                 & R=1    & R=5   & R=10  & R=20  & R=1    & R=5   & R=10  & R=20  \\ \hline
Noise Level: 0\%   &49.3   &76.8   &85.3   &90.0    &58.2        &85.8       &93.4       &97.9      \\
Noise Level: 10\%   &43.4   &70.6   &81.5   &88.9   &52.3        &83.2       &91.4       &97.5       \\
Noise Level: 30\%   &40.0   &67.4   &77.5   &87.0   &51.4        &81.1       &90.5       &96.9       \\
Noise Level: 50\%   &29.8   &60.5   &71.9   &81.5   &44.7        &75.2       &85.6       &95.5       \\
\hline
\end{tabular*}
\end{table}

\subsection{Effectiveness of Feature Averaging}
We also evaluate the performance of our method when different numbers of subsequences are selected in the test phase. Choosing a single subsequence is similar to single-shot case, \textit{i.e.}, the feature of a randomly selected fragment is used for matching. Therefore to be more discriminative, we first randomly select several subsequences from the input human sequence and average the subsequence level human features (based on our recurrent feature aggregation network) over several subsequences. The results are shown in Table~\ref{table3}. As expected, averaging several subsequences significantly enhance the performance because it makes the feature vector more robust to pose/illumination changes for representing a person, \textit{i.e.}, noise is also diluted by this scheme. We also notice that the performance does not have much difference when the number of subsequences is $10$ and $15$. This is because $10$ subsequences are already sufficient to well represent human dynamics.
\begin{table}[t]
\centering
\caption{Performance of our method choosing different number of subsequences}
\label{table3}
\begin{tabular*}{1\textwidth}{@{\extracolsep{\fill} }c|cccc|cccc}
\hline
Dataset                & \multicolumn{4}{c|}{iLIDS-VID} & \multicolumn{4}{c}{PRID 2011} \\ \hline
Rank R                 & R=1    & R=5   & R=10  & R=20  & R=1    & R=5   & R=10  & R=20  \\ \hline
No. of subsequences: 1   &33.1   &60.9   &73.0   &82.9   &45.6        &73.0       &84.7       &94.3       \\
No. of subsequences: 5   &43.9   &70.7   &80.7   &89.2   &52.4        &81.6       &91.7       &97.3       \\
No. of subsequences: 10  &\textbf{49.3}  &\textbf{76.8}  &\textbf{85.3}  &\textbf{90.0}    &\textbf{58.2}   &\textbf{85.8}  &\textbf{93.4}  &97.9       \\
No. of subsequences: 15   &47.8   &72.2   &82.4   &89.7    &57.9   &83.4  &92.8  &\textbf{98.0}       \\
\hline
\end{tabular*}
\end{table}

\subsection{Comparison to the State-of-the-Art}
In this section, our method is compared with the state-of-the-art multi-shot person re-id approaches. We choose three methods for comparison and the results are shown in Table~\ref{table4}. From the results, we see that our method has outperformed the Discriminative Video Ranking (DVR)~\cite{DBLP:journals/corr/WangGZW16} for more than $20\%$ on rank-$1$ matching rate. This is due to the fact that instead of selecting just one discriminative video fragment, we randomly select several subsequences in the video, which makes our feature more discriminative. Our method also achieves better performance than the dictionary learning method DVDL~\cite{DBLP:conf/iccv/KaranamLR15}, which discriminatively trained viewpoint invariant dictionaries. However, DVDL is unable to encode temporal information into the dictionaries. In our case, LSTM effectively solves this problem. Spatio-temporal Fisher vector (STFV3D)~\cite{DBLP:conf/iccv/LiuMZH15} generates a spatio-temporal body-action model that consists of a series of body-action units, which are represented by Fisher vectors built upon low-level features that combines color, texture and gradient information. This representation aligns the spatio-temporal appearance of a person globally and thus displays high discriminativeness when combines with a metric learning method (KISSME~\cite{DBLP:conf/cvpr/KostingerHWRB12}). Compared to this method, our approach is simple and does not need to consider the temporal alignment problem. Even though, we still achieves comparable results to STFV3D+KISSME. In addition, the impressive performance of STFV3D+KISSME is largely due to the effectiveness of the metric learning method. Thus, using no metric learning leads to a large performance drop for this method. On the other hand, our method does not necessarily require a metric learning method to enhance performance. Even using a simple cosine distance metric, it still achieves impressive results on both datasets. This demonstrates that the feature we learned is very discriminative itself.
\begin{table}[t]
\centering
\caption{Performance of our method compared against state-of-the-art methods}

\label{table4}
\begin{tabular*}{1\textwidth}{@{\extracolsep{\fill} }c|cccc|cccc}
\hline
Dataset                & \multicolumn{4}{c|}{iLIDS-VID} & \multicolumn{4}{c}{PRID 2011} \\ \hline
Rank R                 & R=1    & R=5   & R=10  & R=20  & R=1    & R=5   & R=10  & R=20  \\ \hline
3D Hog\&Color+DVR~\cite{DBLP:journals/corr/WangGZW16}   &39.5      &61.1       &71.7       &81.0       &40.0        &71.7       &84.5       &92.2       \\
DVDL~\cite{DBLP:conf/iccv/KaranamLR15}   &25.9        &48.2       &57.3       &68.9       &40.6        &69.7       &77.8       &85.6       \\
STFV3D~\cite{DBLP:conf/iccv/LiuMZH15}  &37.0        &64.3       &77.0       &86.9       &21.6        &46.4       &58.3       &73.8       \\
STFV3D+KISSME~\cite{DBLP:conf/cvpr/KostingerHWRB12}   &44.3   &71.7  &83.7 &\textbf{91.7}  &\textbf{64.1}        &\textbf{87.3}       &89.9       &92.0       \\ \hline
Color\&LBP+\textbf{RFA-Net}+Cosine &44.5  &71.9   &82.0   &90.1    & 54.9    &84.2   &\textbf{93.7}   &\textbf{98.4}  \\
Color\&LBP+\textbf{RFA-Net}+RSVM   &\textbf{49.3}   &\textbf{76.8}   &\textbf{85.3}   &90.0   &58.2 &85.8 &93.4 &97.9\\ \hline
\end{tabular*}
\end{table}

\begin{figure}[t]
  \centering
  \subfigure[]{
    \label{fig:subfig:a} 
    \includegraphics[width=1.2in]{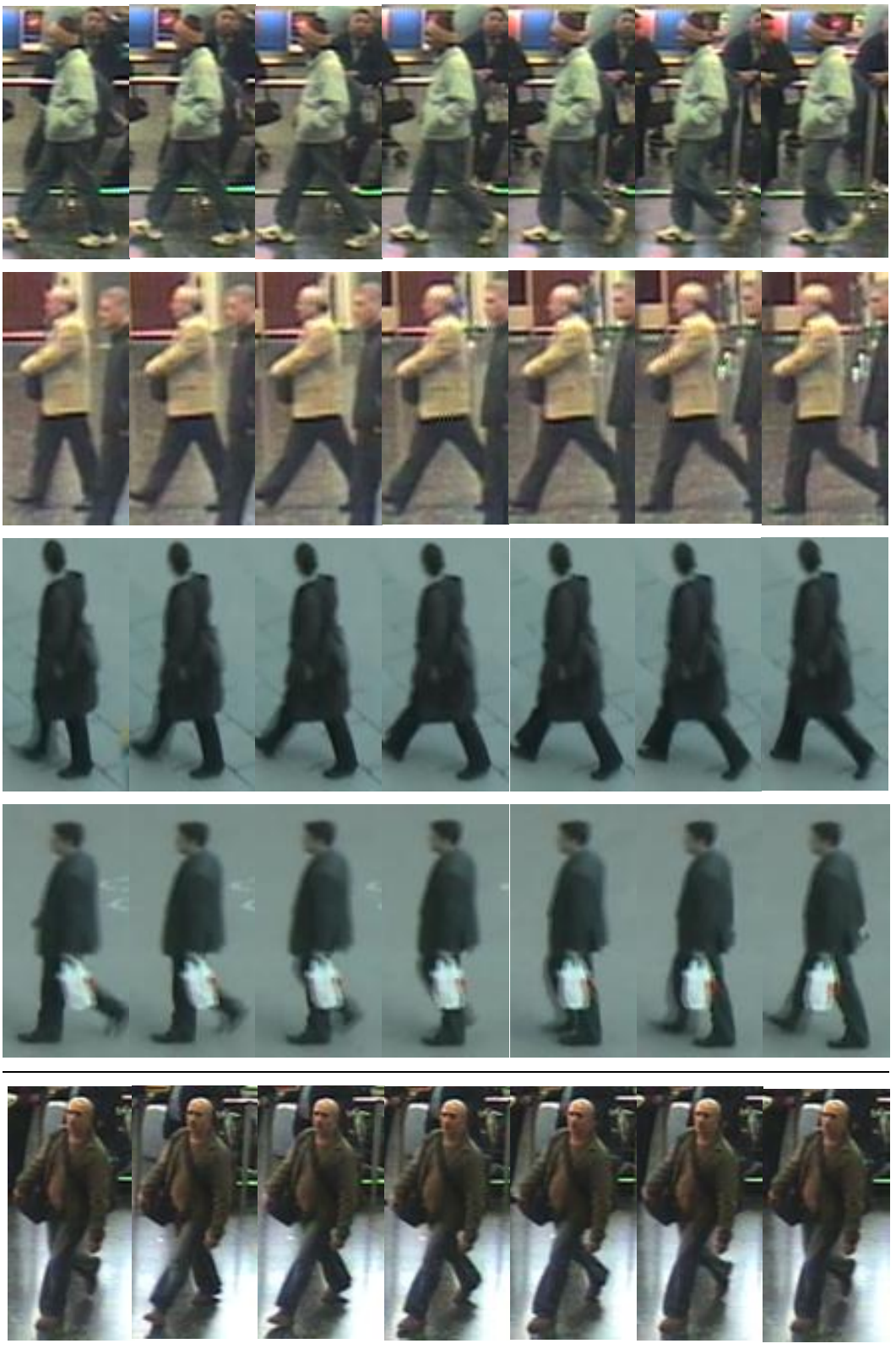}}
  \subfigure[]{
    \label{fig:subfig:b} 
    \includegraphics[width=1.2in]{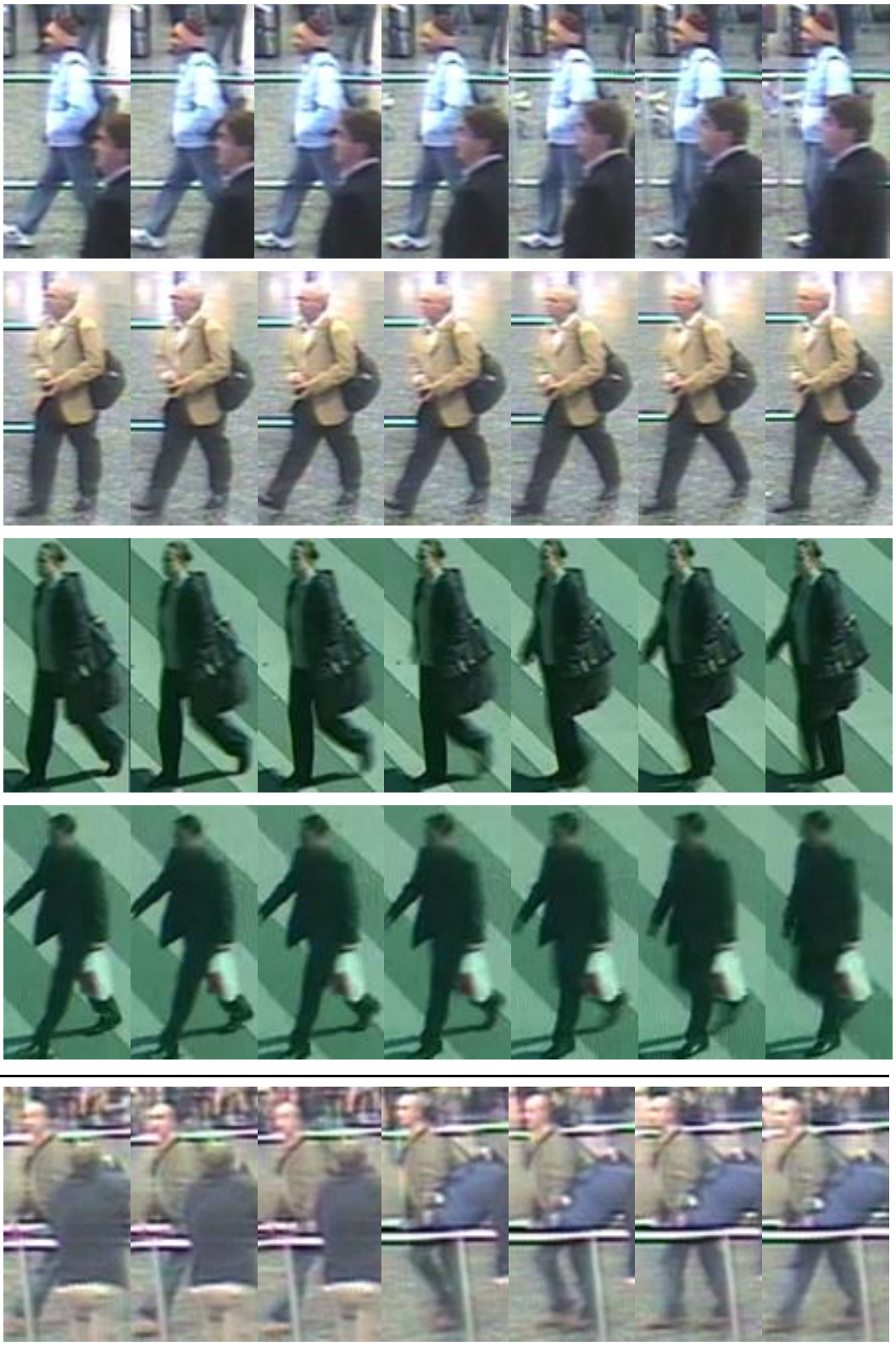}}
    \subfigure[]{
    \label{fig:subfig:a} 
    \includegraphics[width=1.2in]{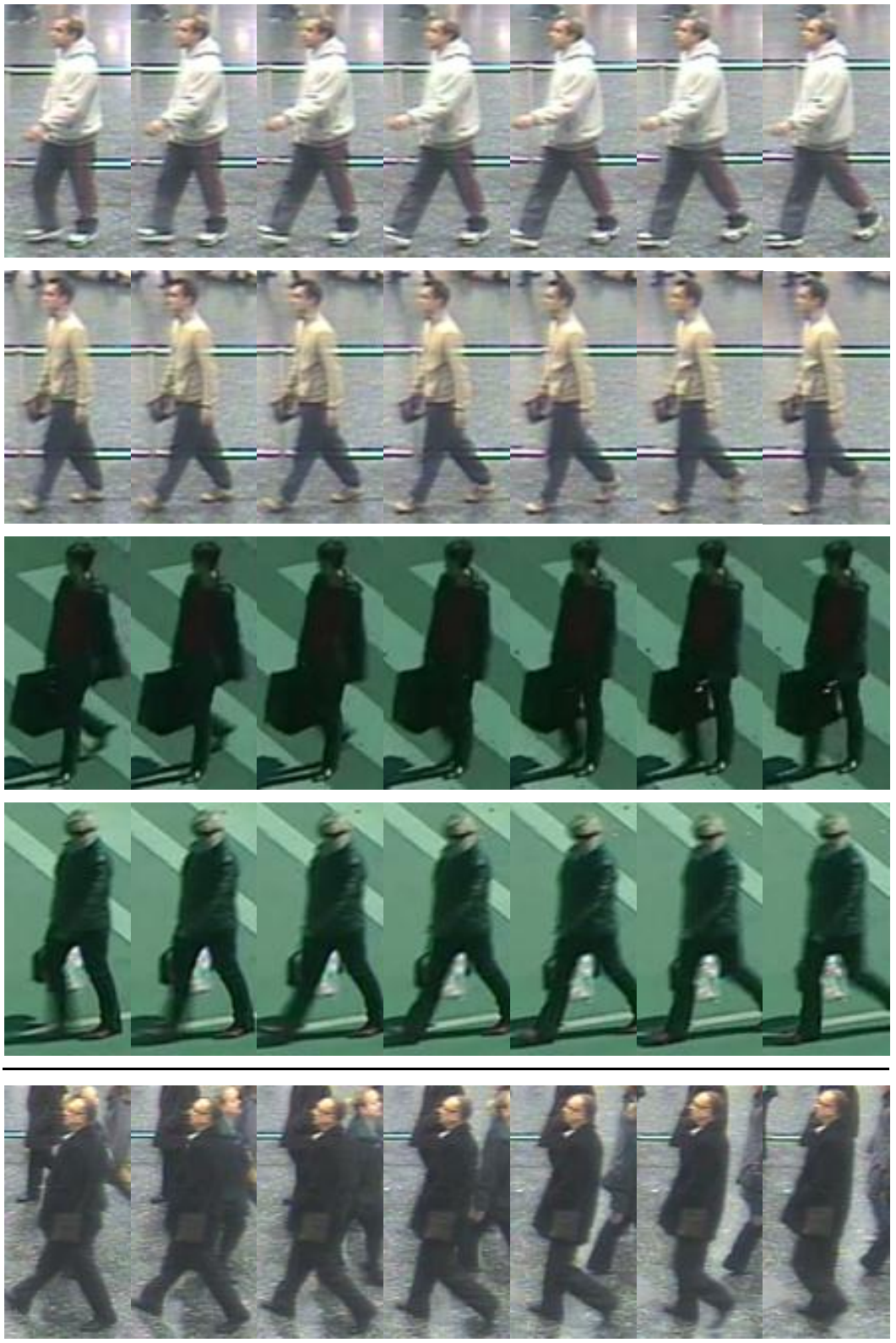}}
  \caption{Matching examples. (a) probe sequence, (b) Correct match, (c) Rank-1 matching sequence. The first four rows correspond to failure examples using features accumulated to the $10$th LSTM node. The bottom row illustrates an example that fails to match the correct sequence when using the features from the first LSTM node, but correctly matches the probe sequence when using the features accumulated to the $10$th LSTM node}
  \label{fig:failure} 

\end{figure}

\subsection{Matching Examples}
In this section, we discuss about some matching examples in our experiments, as shown in Figure~\ref{fig:failure}. Here, four typical failure re-identification examples and a successful example are illustrated, where each row displays one. The first column in the figure consists of probe sequences, the second corresponds to their correct matching pairs and the third one contains the top matching sequence generated by our method. The top two rows show the failure examples from iLIDS-VID datasets, where the first failure is largely due to color inconstancy and occlusion. In the second case, the persons in last two sequences nearly have the same pose and they wear clothes of similar color, which make it difficult for a texture and color based feature to discriminate. In fact, these two sequence have very close similarity scores. The next two failure cases are similar, which are illustrated in the third and fourth row. The matching failure is caused by similar local information such as black briefcase in and white bag in hand, which makes it difficult to identify the correct match. These failure examples demonstrate that both color and local features have their limitations, using or combing more features has great opportunity to enhance the performance of our system.

An interesting example is displayed in the bottom row, where the third column corresponds to the rank-1 matching sequence using the features extracted from the first LSTM node. The failure is due to the occlusion, pose change and color inconsistency in the gallery sequence. However, when features are accumulated to the deeper ($10$th) LSTM nodes, our system correctly matches the probe sequence with the corresponding gallery sequence, which well demonstrate the effectiveness of our feature aggregation method. In addition, notice that all the examples displayed here are also difficult for humans to identify. The performance of our method using simple texture and color features is remarkable.

\section{Conclusions}
In this paper, we proposed a novel recurrent feature aggregation framework for person re-identification. In contrast to existing multi-shot person re-id methods that use complex feature descriptors or design complex matching metric, our method is capable of learning discriminative sequence level representation from simple frame-wise features. Experimental results show that features learned by this recurrent network not only possess high discriminativeness, and also show great robustness to noise. Even using simple texture and color features as input, the performance of the proposed method is better/comparable to the state-of-the-art methods.

\bigskip
\noindent\textbf{Acknowledgements.} The work was supported by State Key Research and Development Program (2016YFB1001003), NSFC (61527804, 61521062, 61502301), STCSM (14XD1402100), the 111 Program (B07022) and China's Thousand Youth Talents Plan.

\bibliographystyle{splncs}
\bibliography{egbib}
\end{document}